\documentclass[10pt,twocolumn,letterpaper]{article}

\usepackage{cvpr}              

\usepackage{graphicx}
\usepackage{amsmath}
\usepackage{amssymb}
\usepackage{booktabs}
\usepackage{multirow}
\usepackage{microtype}
\usepackage{adjustbox}
\usepackage{enumitem}
\usepackage{tabularx}
\usepackage{cuted}
\usepackage[table,xcdraw]{xcolor}
\newcolumntype{Y}{>{\centering\arraybackslash}X}

\newcommand\rurl[1]{%
	\href{http://#1}{\nolinkurl{#1}}%
}

\definecolor{citecolor}{RGB}{9,113,188}
\usepackage[citecolor=citecolor,breaklinks=true,bookmarks=false,colorlinks,pagebackref]{hyperref}

\usepackage[capitalize]{cleveref}
\crefname{section}{Sec.}{Secs.}
\Crefname{section}{Section}{Sections}
\Crefname{table}{Table}{Tables}
\crefname{table}{Tab.}{Tabs.}

\newcommand*{\belowrulesepcolor}[1]{%
	\noalign{%
		\kern-\belowrulesep
		\begingroup
		\color{#1}%
		\hrule height\belowrulesep
		\endgroup
	}%
}
\newcommand*{\aboverulesepcolor}[1]{%
	\noalign{%
		\begingroup
		\color{#1}%
		\hrule height\aboverulesep
		\endgroup
		\kern-\aboverulesep
	}%
}

\def\etal{\emph{et al}\onedot}

\usepackage{amsfonts}
\let\oldsmallfrown\smallfrown
\renewcommand{\smallfrown}[1][0pt]{%
	\mathrel{\raisebox{#1}{$\oldsmallfrown$}}%
}

\def\cvprPaperID{4507} 
\def\confName{CVPR}
\def\confYear{2022}

\begin{document}
	
	\title{Bringing Old Films Back to Life}
	
	\author{Ziyu Wan$^{1}$ \quad \quad Bo Zhang$^{2}$ \quad \quad Dongdong Chen$^3$ \quad \quad Jing Liao$^{1}$\thanks{Corresponding author.} \\
		$^1$City University of Hong Kong \quad $^2$Microsoft Research \quad $^3$Microsoft Cloud + AI
		\\
		{\tt\small \{raywzy,cddlyf\}@gmail.com \quad zhanbo@microsoft.com \quad jingliao@cityu.edu.hk} \\
		\url{http://raywzy.com/Old_Film}
	}
	
	\maketitle

	\def\swseven{0.108\linewidth}
	\begin{strip}
		\vspace{-0.8in}
		
		\setlength\tabcolsep{0.5pt}
		\centering
		\begin{tabular}{cccccc}
			\vspace{-1mm}
			\includegraphics[height=\swseven]{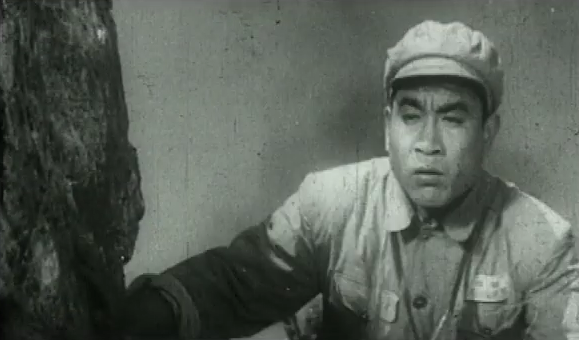}&
			\includegraphics[height=\swseven]{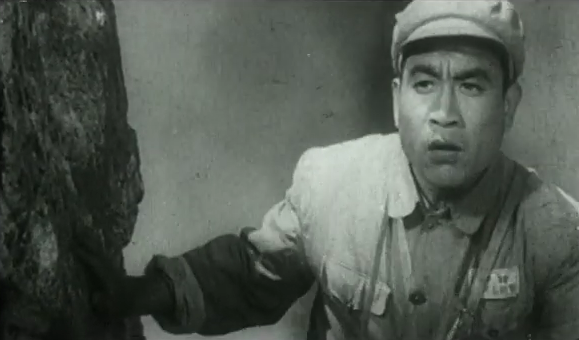}&
			\includegraphics[height=\swseven]{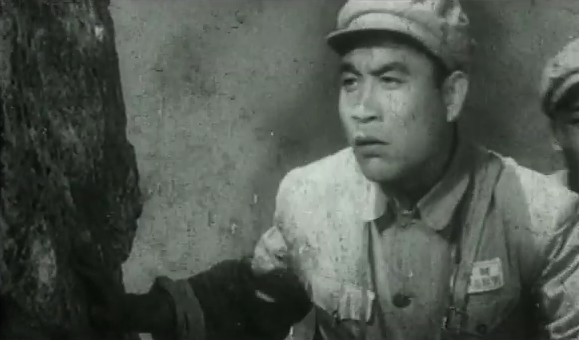}&
			\includegraphics[height=\swseven]{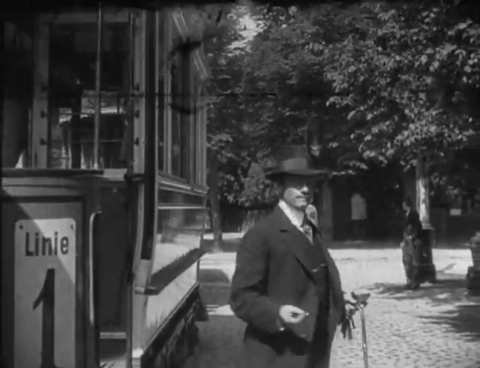}&
			\includegraphics[height=\swseven]{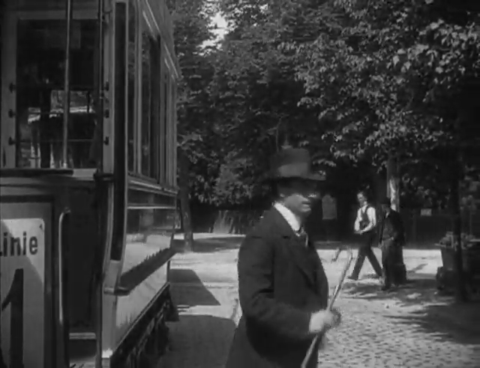}&
			\includegraphics[height=\swseven]{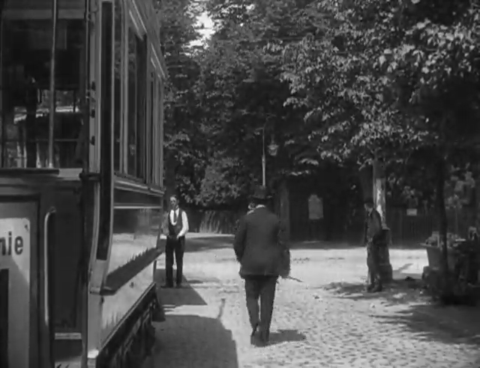}
			\\
			\vspace{-0.5mm}
			\includegraphics[height=\swseven]{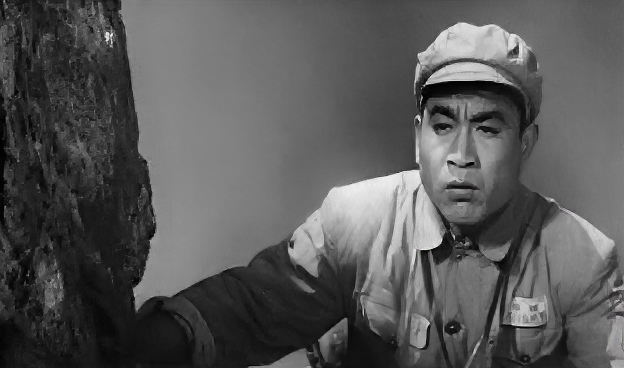}&
			\includegraphics[height=\swseven]{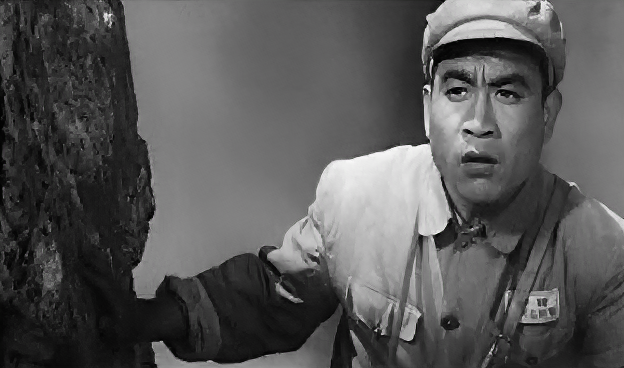}&
			\includegraphics[height=\swseven]{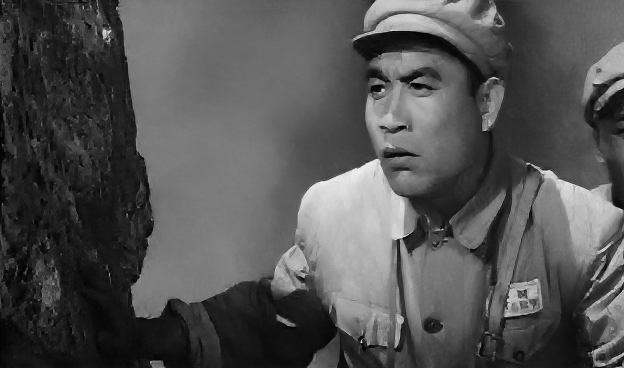}&
			\includegraphics[height=\swseven]{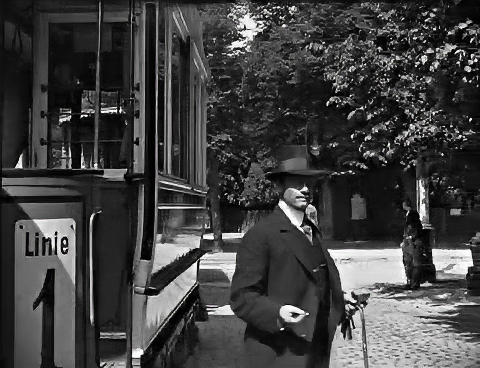}&
			\includegraphics[height=\swseven]{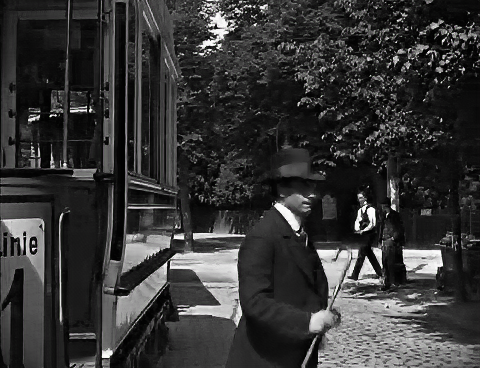}&
			\includegraphics[height=\swseven]{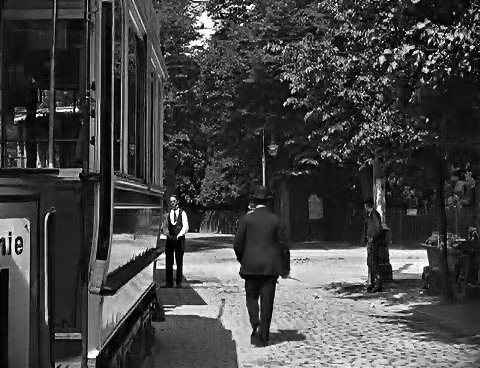}
			\\
			\vspace{-0.5mm} 
			\includegraphics[height=\swseven]{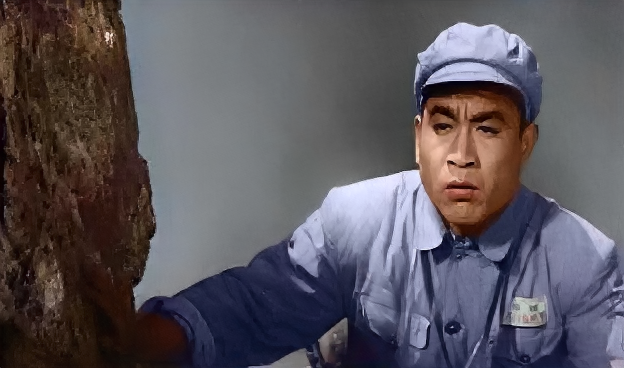}&
			\includegraphics[height=\swseven]{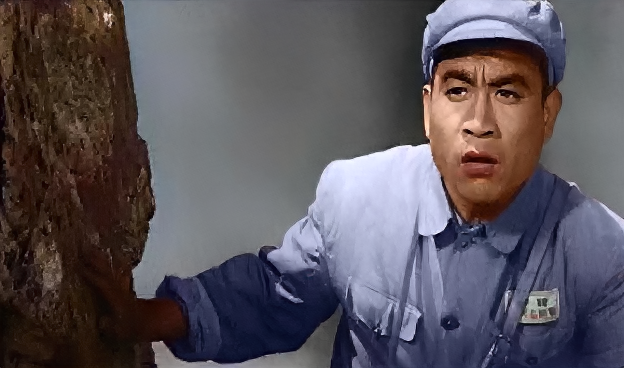}&
			\includegraphics[height=\swseven]{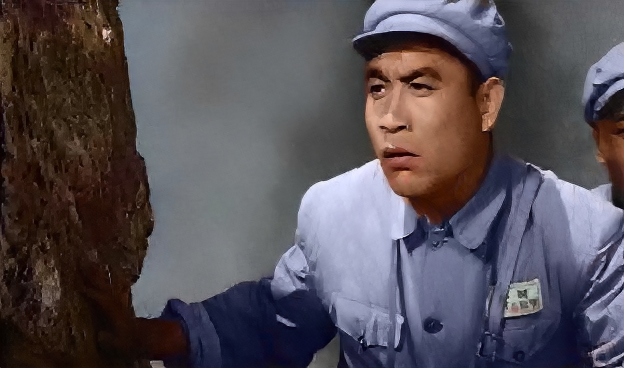}&
			\includegraphics[height=\swseven]{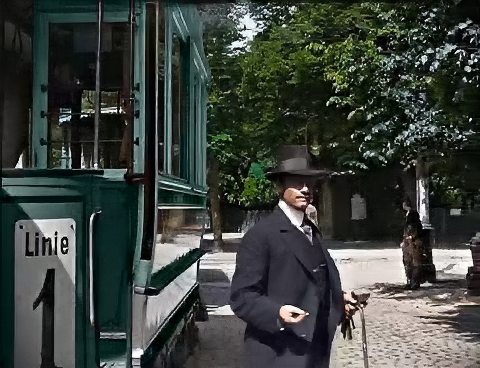}&
			\includegraphics[height=\swseven]{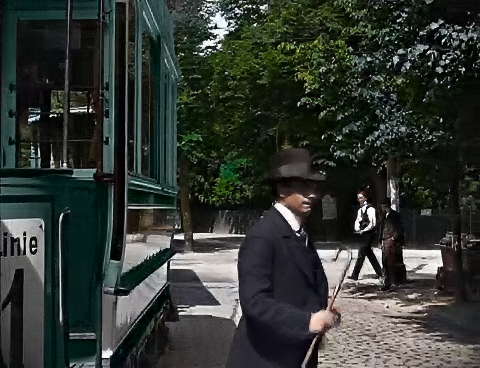}&
			\includegraphics[height=\swseven]{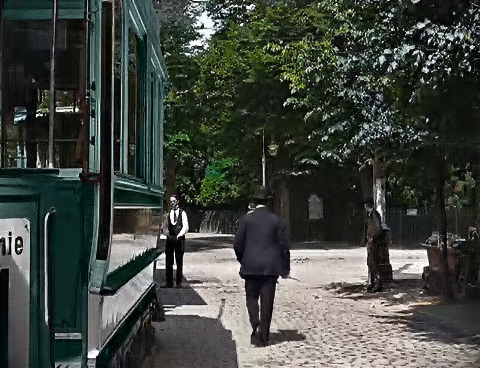}
			
		\end{tabular}
		\vspace{-5pt}
		\captionof{figure}{{Real-world old films restored by our method.} First row: original video frames. Second row: the restored frames. Third row: the restored frames after further colorization. The quality of old films is considerably enhanced after restoration.}
		\label{fig:teaser}
		\vspace{-0.2em}
	\end{strip}
	\begin{abstract}
		\vspace{-1.0em}
		We present a learning-based framework, recurrent transformer network (RTN), to restore heavily degraded old films. Instead of performing frame-wise restoration, our method is based on the hidden knowledge learned from adjacent frames that contain abundant information about the occlusion, which is beneficial to restore challenging artifacts of each frame while ensuring temporal coherency. Moreover, contrasting the representation of the current frame and the hidden knowledge makes it possible to infer the scratch position in an unsupervised manner, and such defect localization generalizes well to real-world degradations. To better resolve mixed degradation and compensate for the flow estimation error during frame alignment, we propose to leverage more expressive transformer blocks for spatial restoration. Experiments on both synthetic dataset and real-world old films demonstrate the significant superiority of the proposed RTN over existing solutions. In addition, the same framework can effectively propagate the color from keyframes to the whole video, ultimately yielding compelling restored films. The implementation and model will be released at \href{https://github.com/raywzy/Bringing-Old-Films-Back-to-Life}{https://github.com/raywzy/Bringing-Old-Films-Back-to-Life}.

	\end{abstract}
	
	\section{Introduction}
	\label{sec:intro}
	Old film classics have the lasting power to strike the resonance and fantasies of audiences today. Unfortunately, many of them have become less popular because people are no longer used to the low resolution and disturbing artifacts caused by the photographic film aging. Film restoration techniques have been developed to bring these old films back to life, which nonetheless takes painstaking efforts. The restoration nowadays is conducted digitally, where the artists meticulously examine each frame, manually retouch the blemishes, fix up the flickering and finally perform the colorization frame by frame, so repairing the entire old film brings insurmountable expenses. Hence, people desire an algorithm that performs all of these tedious tasks automatically such that old films can be revived in modern looking.
	
	Old films typically suffer from mixed degradations, which, to the best of our knowledge, only a few works aim to solve. While one can sequentially apply dedicated models for restoration, the models designed for specific tasks cannot generalize well to real-world degradations. Recently, higher-order degradation models~\cite{zhang2021designing,wang2021real} have been proposed to characterize the real-world degradations, yet these works mainly consider the photometric degradations, such as blurriness and noises, rather than the structured defects (\eg, scratches, cracks, etc.) that obstruct the most in old films. Closely related to our work, ~\cite{wan2020bringing} attempts to address complex degradations in vintage photos, yet its frame-wise processing on old films does not yield temporally consistent results. The DeepRemaster~\cite{iizuka2019deepremaster}, in comparison, targets video restoration as well as colorization, yet this work cannot sufficiently leverage the temporal information with explicit frame alignment and the spatial information with long-range correlations, thus unable to fix up large cracked areas.
	
\begin{figure}[!t]
		\begin{center}
			\includegraphics[width=1.0\linewidth]{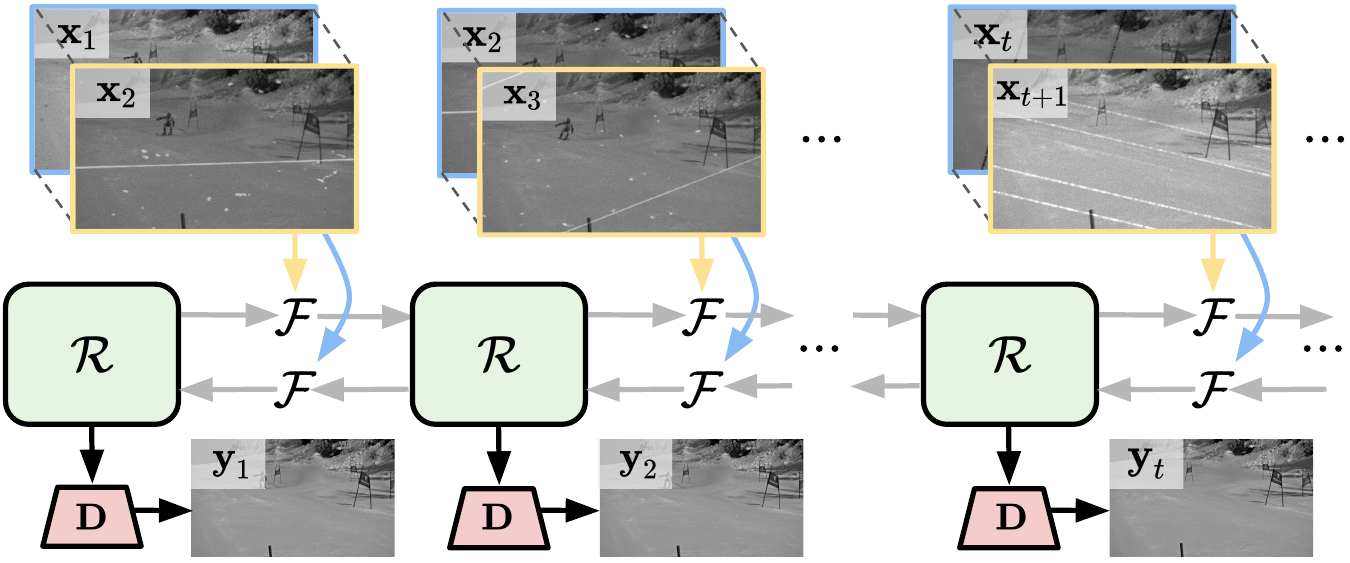}
			\vspace{-1.4em}
			\caption{{Pipeline overview.} Our method follows a bi-directional RNN architecture. $\mathcal{R}$: Spatial restoration. $\mathcal{F}$: Feature aggregation. $\mathbf{D}$: Pixel reconstruction decoder.}
			\label{fig:overall_pipeline}
		\end{center}
		\vspace{-2.5em}
	\end{figure}

	In this work, we seek to unify the entire film restoration tasks with a single framework in which we conduct spatio-temporal restoration. The key insight is that most degradations in old films, especially structured defects, are temporally variant, \ie, the structured defects occluded in one frame may reveal its content in successive frames. Therefore, we propose to repair the degradations by leveraging the spatio-temporal context rather than relying on the hallucination. Specifically, we propose a bi-directional recurrent network (Figure.~\ref{fig:overall_pipeline}) which aggregates the knowledge of the scene across adjacent frames, effectively reducing the film flickering. The hidden state of the recurrent module embeds the representation of the scene content. After the alignment, the restoration for a specific frame fuses such hidden representation as it offers useful knowledge of the film content underlying the defects. Such a recurrent scheme brings three-fold benefits. First, the film degradations, no matter how severe they are, can be fully restored as long as the information is well-preserved in other frames. Second, the explicit maintenance of the hidden knowledge ensures that the restoration for frames is temporally consistent in a long period. More importantly, the structured defects can be localized in an unsupervised manner because these areas show a larger discrepancy between the representation of the current frame and the hidden state. As opposed to~\cite{wan2020bringing} that requires a defect segmentation network, such defect localization is more generalizable to real-world old film degradations.
	
	Spatially, we need a module to account for the slight mismatch during the frame alignment. As such, we propose to leverage the Swin Transformer~\cite{Liu_2021_ICCV} --- even if the hidden representation is not accurately aligned, the interaction of the corresponding pixels can still be modeled through self-attention. Indeed, we observe more stabilized training of the recurrent module due to the use of attention. Besides, thanks to the superior expressivity, the transformer blocks offer improved restoration ability for mixed degradations which would be hard to resolve using a specialized ConvNet~\cite{yu2018crafting,suganuma2019attention}. Thus, the proposed network makes the best of the recurrent module and transformers: the memorization nature of recurrent modules benefits the temporal coherency whereas the long-range modeling capability of transformers helps the spatial restoration, which significantly outperforms strong baselines on synthetic datasets, and yields unprecedented quality when restoring real old films.

	Moreover, we show that the same framework can be easily adapted for film colorization as well. We follow the colorization pipeline favored by artists that only a few keyframes undergo manual colorization, whose color is then propagated to the rest frames. Our method performs favorably over the leading colorization methods, effectively propagating the color from keyframes to the whole video. As shown in Figure.~\ref{fig:teaser}, our method unifies the restoration and colorization, and demonstrates the capability of reviving the old films as if they were captured by yesterday.

	\begin{figure*}[!t]
		\begin{center}
			\includegraphics[width=1.0\linewidth]{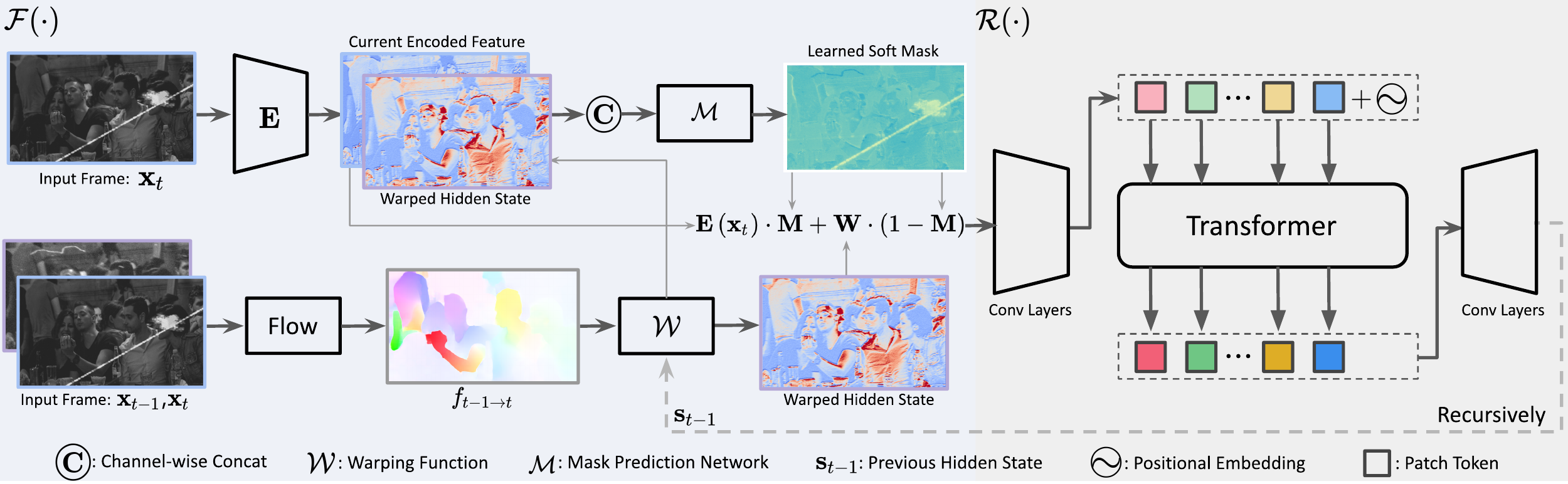}
			\vspace{-1.4em}
			\caption{{The framework of temporal aggregation module $\mathcal{F}$ and spatial restoration transformer $\mathcal{R}$ in once recurrent forward propagation.} Backward propagation follows the same paradigm.}
			\label{fig:recurrent_pipeline}
		\end{center}
		\vspace{-2.0em}
	\end{figure*}
	
	\section{Related Works}
	
	\noindent\textbf{Image Restoration} ~~ Traditional methods~\cite{buades2005non,elad2006image,weiss2007makes,babacan2008total} mostly use model-based optimization to restore degraded images, where various image priors are imposed to find suitable solutions. Recently, CNN-based methods have shown remarkable capability for restoration through learning mappings between low-quality and high-quality paired images. However, these works only focus on a single type of degradation like denoising~\cite{zhang2017beyond,zhang2018ffdnet,mao2016image}, super-resolution~\cite{kim2016accurate,ledig2017photo,zhang2018residual,wang2018esrgan}, inpainting~\cite{yu2019free,liu2021pd,Wan_2021_ICCV}, colorization\cite{he2018deep}, \etc, whereas real-world images are plagued with a compound of degradations. Although some methods~\cite{zhang2021designing,wang2021real,liang2021swinir} start to improve the results of real settings via constructing degradation models, they mainly focus on unstructured defects and leave the scratches or dirt on the old medium unfixed. Considering this issue, Wan~\etal~\cite{wan2020bringing, wan2020old} propose a comprehensive solution for old photo restoration by restoring the global and local artifacts in the latent space. Although impressive results can be achieved, they cannot handle the artifacts for dynamic scenes well.
	
	\noindent\textbf{Video Restoration} ~~ 
	Most existing methods explicitly estimate the dense correspondence among input
	frames, then reconstruct the clean target frame via CNNs for video denoising\cite{claus2019videnn,tassano2020fastdvdnet}, video deblurring~\cite{su2017deep,zhou2019spatio} or video super-resolution~\cite{wang2019edvr,haris2019recurrent,chan2021basicvsr}. However, they barely consider the real-world mixed degradations and only have limited generation capability. Another research line, video inpainting~\cite{xu2019deep,kim2019deep,gao2020flow}, tries to synthesize plausible contents for the missing regions of each frame conditioned on global information. Yet, these works assume a fixed inpainting mask prescribed by users, whereas the defect areas in old films are typically unknown and temporally varying. 
	
	\noindent\textbf{Old Film Restoration} ~~ Traditional approaches~\cite{giakoumis2005digital,hongying2009efficient,kim2006efficient,saito1999image} focus on removing the structured artifacts with a detection network followed by an inpainting pipeline, but these works rely on hand-crafted features without a semantic understanding of the video content, which limits their inpainting effects. Moreover, focusing on structured defects while ignoring photometric degradations (\eg, blurriness and noises) makes their overall restoration results less appealing. Subsequently, Iizuka~\etal propose DeepRemaster~\cite{iizuka2019deepremaster}, a fully 3D convolutional method, to restore the old films. This method works well on restoring the artifacts of synthetic videos, but it fails to generalize well on restoring real-old films, especially the ones with large cracked areas because it lacks special designs for complex mixed degradation and fails to sufficiently leverage the long-term temporal information from adjacent frames with 3D convolution.

	\section{Method}
	
	Let $\mathbf{x}_{1}^{T} \equiv\left\{\mathbf{x}_{1}, \mathbf{x}_{2}, \ldots, \mathbf{x}_{T}\right\}$ be a sequence of old film frames, where $T$ is the video length. Our target is to train a deep neural network to automatically restore spatial-temporal deterioration. To comprehensively mitigate the issues of old films, like illuminance flicker, physical structural pollution, or quality degeneration, we propose recurrent transformer networks (RTN), whose details are elaborated in Sec.~\ref{sec3.1}. Since the training of RTN requires supervision, we subsequently introduce a data simulation method in Sec.~\ref{sec3.2}, which could convert consecutive video frames acquired by modern cameras to corresponding degraded versions. In Sec.~\ref{sec3.3}, we further extend the proposed framework to reference-based video colorization.  Lastly,  we introduce how to optimize RTN in Sec.~\ref{sec3.4}. 
	
	\subsection{Proposed Framework}\label{sec3.1}
	
	\noindent\textbf{Temporal Recurrent Network} ~~ Most of the time, the frames of old films are plagued with severe quality degradation due to the improper storage environment of the film material and the abrasion caused by mechanical protruding parts of old-fashioned projectors. More seriously, sometimes several frames are totally damaged where only fuzzy structures remain. Hence, it is necessary to leverage the long-term temporal clues to remove the occurred degradations and then render reasonable contents.
	
	Towards this goal, we propose a recurrent architecture for the temporal modeling of old films because of the huge benefits of learned hidden representation and its powerful long-term propagation ability.  Specifically, for timestamp $t$, we estimate the optical flows $f_{t-1 \rightarrow t}$ between two input frames $\mathbf{x}_{t-1}$ and $\mathbf{x}_{t}$. Then the previous hidden state $\mathbf{s}_{t-1}$ could be aligned to time $t$ by warping function $\mathcal{W}$. Meanwhile, we leverage a convolutional encoder $\mathbf{E}$ to project input frame $\mathbf{x}_{t}$ into feature map $\mathbf{E(\mathbf{x}_{t})}$ which shares the same spatial dimension with propagated state, then the newly restored state $\mathbf{s}_t$ could be obtained as follows,
	\begin{equation}
	\mathbf{s}_t=\mathcal{R}_{\uparrow} \circ \mathcal{F}_{\uparrow}(\mathbf{E}(\mathbf{x}_t),\mathcal{W}(\mathbf{s}_{t-1},f_{t-1 \rightarrow t})),
	\end{equation}
	where $\mathcal{F}$ aggregates the states between history and current feature, and $\mathcal{R}$ further recovers and boosts the hidden representation, which will be described in the next section.
	
	Due to the irregular exposure time of early movie cameras, the global brightness or colors may vary from one frame to the next, which is called the \textit{flicker} phenomenon. The flicker is more challenging in our case because the content of old film itself is flickering whereas the flicker in typical video restoration tasks is mainly caused by frame-wise processing. Therefore, we need to aggregate the information from frames in a long period. To ensure that bidirectional information contributes equally, we also learn the backward hidden state $\tilde{\mathbf{s}}_t$ by considering the knowledge of future frames, \begin{equation}
	\tilde{\mathbf{s}}_t=\mathcal{R}_{\downarrow} \circ \mathcal{F}_{\downarrow}(\mathbf{E}(\mathbf{x}_t),\mathcal{W}(\tilde{\mathbf{s}}_{t+1},f_{t+1 \rightarrow t})).
	\end{equation}
	Combining the bidirectional states ${\mathbf{s}}_t$ and $\tilde{\mathbf{s}}_t$, we reconstruct the final pixel-level output via a convolutional decoder. \begin{equation}
	\mathbf{y}_{t} = \mathbf{D}(\mathbf{s}_t \smallfrown[4pt] \tilde{\mathbf{s}}_t ) \in \mathbb{R}^{H \times W },
	\end{equation}
	where $\smallfrown[4pt]$ denotes the concatenation operation along the channel dimension. To this end, we have known how to effectively propagate the temporal information by an RNN, and the overall system pipeline is shown in Figure.~\ref{fig:overall_pipeline}.

	\noindent\textbf{Spatial Transformer Restoration} ~~ After blending the hidden state and current feature, we need another spatial network to perform the frame-wise restoration. CNN has been a de-facto standard to most video processing tasks due to its efficient locality and excellent pattern restoration ability. However, the same locality property does not apply to old film restoration considering the following several aspects: 1) Old films contain many non-spatial homogeneous degradations like dust, dirt, or scratches, which require the network to explore long-range context information for plausible inpainting. 2) Since old films always deteriorate excessively, the estimated position-wise correspondence between frames may contain errors. If we still use CNN for spatial restoration, the consequent results are extremely unstable training, which has been verified in our experiments. In such cases, we need a mechanism to compensate for the inevitable estimation errors of optical flow.
	
	To alleviate the above-mentioned issues, we propose to employ a powerful transformer network for spatial restoration. The problem is the quadratic computational cost of transformer architecture, which makes it difficult to model high-dimensional data. To efficiently process old films for high-resolution images, we adopt two modifications. Let's denote the fused representation derived from $\mathcal{F}(\cdot)$ as $\mathbf{h}\in \mathbb{R}^{H \times W \times C}$.  First, we use strided convolutions to downsample the spatial resolution by a factor of 2. Second, instead of directly modeling all the correlations among $\frac{HW}{4}$ tokens, we follow Swin Transformer~\cite{Liu_2021_ICCV} and employ the window-based attention and shifted strategy to improve the efficiency further. More specifically, in each layer of transformer, given one feature representation $\mathbf{z}\in \mathbb{R}^{\frac{H}{2} \times \frac{W}{2} \times C}$, we partition it into several non-overlapping $M \times M$ local windows. In each local window, we project the tokens into queries $\mathbf{Q}$, keys $\mathbf{K}$ and values $\mathbf{V}$ respectively through MLPs. Then the attention could be computed as follows,
	\begin{equation}
	\operatorname{Att}(\mathbf{Q}, \mathbf{K}, \mathbf{V})=f(\mathbf{Q}\mathbf{K}^{\top} / \sqrt{d}+\mathbf{B})\mathbf{V},
	\end{equation}
	
	where $\mathbf{Q}, \mathbf{K}, \mathbf{V} \in \mathbb{R}^{M^2 \times d}$, $f(\cdot)$ is the softmax operation and $\mathbf{B}$ is the learnable relative positional embedding. Despite higher efficiency, the local attention would inevitably hurt the global modeling capability. To solve this problem, we also involve the cross-window connections with a cyclic shift. With stacking the consecutive transformer block, the model is able to perceive global context information for better degradation restoration and compensation of flow errors.

	\noindent\textbf{Learnable Guided Mask} ~~ Intuitively, to restore the contaminants lying on frames, we could first localize the exact mask positions of such structured defects and then perform video inpainting. This scheme sounds reasonable, but actually infeasible here due to the following two aspects: 1) Similar to ~\cite{wan2020bringing}, we also tried to train an extra detection model to distinguish the structured degradations. Nonetheless, the generalization capability of this detection model is extremely deficient, especially for real-world data. The underlying reason is that, unlike single images, the diversities and variations of old film contaminants are larger. Although mixing some real-world paired data in training may alleviate this problem, the labeling cost is too large to afford, especially for video data. 2) Instead of losing all pixel information, by contrast, the dirt in one frame may be transparent, whose preserved contents may guide us for better restoration.
	
	Motivated by that the contaminants mostly have large motion, shape and material variations across different frames, our solution is to learn a soft guided mask by contrasting the hidden state and current frame features in an unsupervised manner. More specifically, as shown in Figure.~\ref{fig:recurrent_pipeline}, given warped clean state $\mathbf{W} = \mathcal{W}(\mathbf{s}_{t-1},f_{t-1 \rightarrow t})$, which has aggregated all available information from frame $0$ to frame $t-1$, we employ a shallow network $\mathcal{M}$ to regress the blending mask $\mathbf{M}$ conditioned on current features $\mathbf{E}(\mathbf{x}_t)$,
	\vspace{-0.5em}
	\begin{equation}
	\mathbf{M} = \mathcal{M}(\mathbf{E}(\mathbf{x}_t) \smallfrown[4pt] \mathbf{W} ) \in \mathbb{R}^{H \times W \times 1 },
	\vspace{-0.5em}
	\end{equation}
	where $\mathcal{M}(\cdot)$ is composed of several convolutions and sigmoid function. With the learned soft mask, we aggregate the temporal priors and current frames by the following method,
	\vspace{-0.5em}
	\begin{equation}
	\mathcal{F}\left(\mathbf{E}\left(\mathbf{x}_{t}\right), \mathbf{W}\right) =\mathbf{E}\left(\mathbf{x}_{t}\right) \cdot \mathbf{M}+\mathbf{W} \cdot(1-\mathbf{M}).
	\end{equation}
	
	\subsection{Video Degradation Model}\label{sec3.2}

	To train the RTN, designing a realistic video degradation model to generate paired data is essential. We achieve this target by involving the following degradation procedures.
	
	\noindent\textbf{Contaminant Blending} ~~ To model the scratches, dirt or dust of old films, we collect 1k+ texture templates from the Internet, which are further augmented with random rotation, local cropping, contrast change, and morphological operations. Then we use \textit{addition}, \textit{subtract} and \textit{multiply} blending modes with various levels of \textit{opacity}$\in [0.6,1.0]$ to combine the scratch textures with the natural frames.
	
	\noindent\textbf{Quality Degradation} ~~ As mentioned before, most old films suffer from blurring, noise and unsharpness due to long-term storage and usage. Hence, we also downgrade the overall video qualities from these aspects. 1)  Gaussian noise and speckle noise with $\sigma \in[5,50]$. 2) Isotropic and anisotropic Gaussian blur kernels with covariance matrix $\boldsymbol{\Sigma}=\boldsymbol{R}\left[\begin{array}{cc}
	\sigma_{1}^{2} & 0 \\
	0 & \sigma_{2}^{2}
	\end{array}\right] \boldsymbol{R}^{T}$, where $\boldsymbol{R}$ denotes the rotation matrix whose rotation angle $\theta \in [0,\pi]$, and $\sigma_{1}, \sigma_{2}\in (0,1)  $ control the  standard deviation of two principal axes respectively.  3) Random downsampling and upsampling with different interpolation methods. 4) JPEG compression whose
	level is in the range of $[40,100]$. 5) Random color jitter through adjusting the \textit{brightness}$\in [0.8,1.2]$ and \textit{contrast}$\in [0.9,1.0]$.
	
	\noindent\textbf{Temporal Frames Rendering} ~~ Unlike the degradation of a single photo, where we could inject these defects randomly, one observation of old films is a similar degradation pattern is shared across frames, which makes sense since most of the frames of one video potentially experience the same mechanical abrasion. Hence, we first define a set of template parameters for each video. Then we further apply predefined parameters with slight perturbations on consecutive temporal frames to achieve more realistic rendering. The synthetic data example is shown in Figure.~\ref{fig:synthetic_show}.
	
	\def\swthreee{0.33\linewidth}
	\begin{figure}
		\renewcommand{\tabcolsep}{0.5pt}
		\begin{center}
			\begin{tabular}{ccc}
				\vspace{-0.5mm}\includegraphics[width=\swthreee]{images/synthetic_data/input/00003.png}&
				\includegraphics[width=\swthreee]{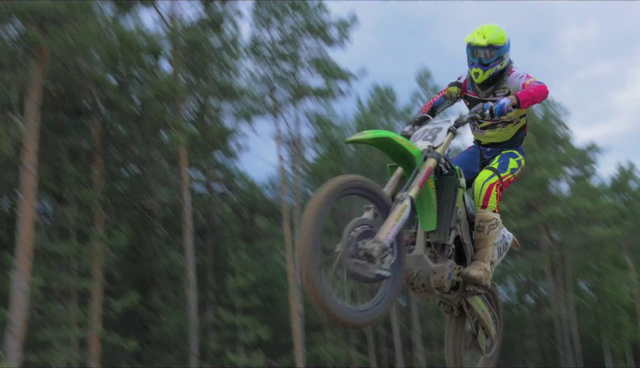}&
				\includegraphics[width=\swthreee]{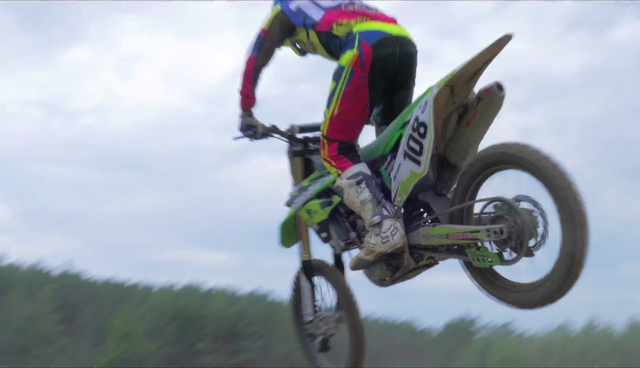} \\
				\includegraphics[width=\swthreee]{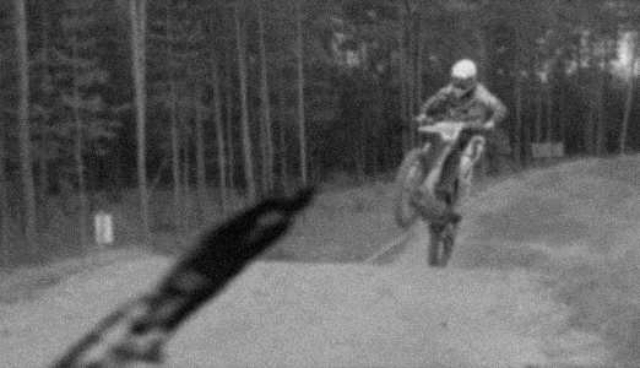}&
				\includegraphics[width=\swthreee]{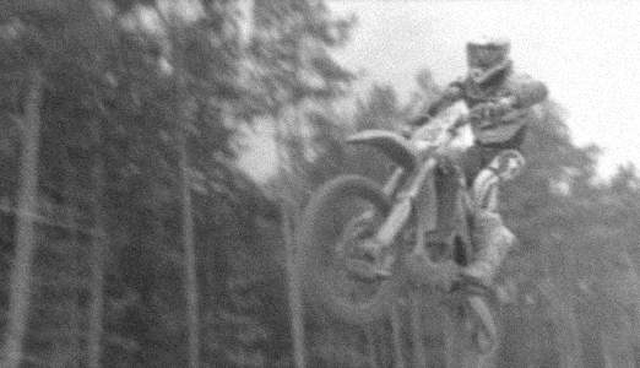}&
				\includegraphics[width=\swthreee]{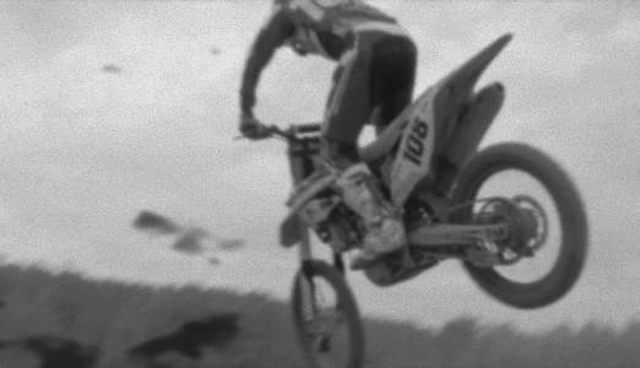} \\
			\end{tabular}
		\end{center}
		\vspace{-1.4em}
		\caption{{Examples of rendered video frames with degradation.}}
		\label{fig:synthetic_show}
		\vspace{-2.0em}
	\end{figure}

	\subsection{Video Colorization}\label{sec3.3}
	
	To make the restored old films more vivid, an effective solution is to colorize the videos further. Due to the ill-posed property of colorization, the automatic algorithms always struggle to generate appealing results~\cite{lei2019fully}. A more practical scenario is, given one frame which has been colorized, how to propagate the known information to all the rest frames, meanwhile preserving the spatial-temporal coherence. Thus we further extend our RTN to the video colorization task under this setting, with only several minor modifications. First, we adjust the input color space from RGB to LAB, and only predict the chrominance channels given black-white frames. Second, to further prevent the color fading of long-term propagation, following~\cite{he2018deep,zhang2019deep} we search the correspondence between two frames via comparing the semantic similarity to get a coarse colorization result. Then the coarse $AB$ channel will be concatenated with gray input and sent into the RTN. With the temporal propagation and spatial refinement capability of RTN, the reference-based colorization could achieve more stable performance. 
	\subsection{Learning}\label{sec3.4}
	
	The ultimate target of our method is to generate temporal-coherent, higher-quality and visual-pleasant videos given consecutive frames from old films. To accomplish these objectives, we apply the following losses.
	
	\noindent\textbf{L1 Loss}  ~~ The pixel-wise reconstruction loss $\mathcal{L}_{{1}}$ penalizes the difference between restored results and GT frames. Specifically, let $\hat{\mathbf{y}_{t}}$ denote target clean video frame and $\mathbf{y}_{t}$ denote the generated frame by our method at time $t$, we have
	\begin{equation}
	\mathcal{L}_{{1}}=\frac{1}{T} \sum_{t=1}^{T} \left\|\mathbf{y}_{t}-\hat{\mathbf{y}_{t}}\right\|_{1}.
	\end{equation}

	\noindent\textbf{Perceptual Loss} ~~  To further improve the visual quality of the restored frames, we also adopt the perceptual loss~\cite{johnson2016perceptual}:
	
	\begin{equation}
	\mathcal{L}_{perc}=\frac{1}{T}\sum_{t=1}^{T} \sum_{p \in P}\omega_{p} \left\|\Phi_{p}^{\mathbf{y}_{t}}-\Phi_{p}^{\hat{\mathbf{y}_{t}}}\right\|,
	\end{equation}
	where $\Phi_{p}^{\mathbf{y}_{t}}$ and $\Phi_{p}^{\hat{\mathbf{y}_{t}}}$ denote the activation from the $p^{th}$ layer of pretrained VGG19 network. $P$ denotes the selected set ($relu2\_2$ to $relu5\_2$) of layers to compute the perceptual loss, and $\omega_{p}$ controls the importance of different layers. The loss is accumulated over all frames in the generated video.

	\noindent\textbf{Spatial-Temporal Adversarial Loss}  ~~  Adversarial training, whose main principle behind it is to process a zero-sum game between a generator and a discriminator, has shown promising results in many tasks~\cite{goodfellow2014generative,wan2020bringing}. For the old film restoration task, we adopt the Temporal-PatchGAN as suggested in~\cite{chang2019free} to enhance both perceptual quality and spatial-temporal coherence. Specifically, we employ a discriminator $D$ composed of 3D convolutions to distinguish each spatial-temporal feature as real or fake by hinge loss,
	\begin{equation}
	\mathcal{L}_{\mathrm{D}}=\mathbb{E}_{\mathbf{y} \sim Y}[ \mathtt{ReLU}({1}-D(\mathbf{y}))]+\mathbb{E}_{\hat{\mathbf{y}} \sim \hat{Y}}\left[\mathtt{ReLU}\left( {1}+D(\hat{\mathbf{y}})\right)\right],
	\end{equation}
	where $Y$ and $\hat{Y}$ are the restored frames of RTN and clean video frames respectively. Then we use RTN to fool the discriminator by:
	\begin{equation}
	\mathcal{L}_{\mathrm{G}}=-\mathbb{E}_{\mathbf{y} \sim Y}[D(\mathbf{y})].
	\end{equation}

	\noindent\textbf{Full Objective}  ~~  Combining all the above losses, the
	the overall objective we aim to optimize is
	\begin{equation}
	\mathcal{L}_{total} = \lambda_{1}\mathcal{L}_{{1}}+\lambda_{p} \mathcal{L}_{perc} + \lambda_{G} \mathcal{L}_{G}.    
	\end{equation}
	We empirically set the weights for different losses as: $\lambda_{1}=1.0$, $\lambda_{p}=1.0$ and $\lambda_{G}=0.01$. For the video colorization task, we calculate the $\mathcal{L}_1$ loss in LAB color space and use a differentiable operator to convert LAB space to RGB space to compute the perceptual loss and adversarial loss.

	\section{Experiments}

	\subsection{Implementation}
	We train the proposed RTN for 20 epochs using the ADAM optimizer~\cite{kingma2014adam} with $\left(\beta_{1}, \beta_{2}\right)=(0.9,0.99)$. The learning rate is set to $2\text{e-}4$ for both generators and discriminators in the first 20 epochs, with linear decay to zero thereafter. We employ the off-the-shelf method RAFT~\cite{teed2020raft} for flow estimation, whose parameters are fixed during the first 5 epochs, and then jointly optimized with other modules together using $\text{lr} = 2.5\text{e-}5$. During optimization, we randomly crop $256$ patches from REDS~\cite{Nah_2019_CVPR_Workshops_REDS} dataset and apply the proposed video degradation model on the fly. The batch size is set to $4$ and the whole training takes $\sim 2$ days on $4$ RTX 2080Ti GPUs.
	
	\subsection{Setup}
	
	\noindent\textbf{Baseline Methods} ~~ We conduct comprehensive experiments by comparing RTN with the following approaches:
	\begin{itemize}[leftmargin=*]
		\itemsep0em 
		\item \textit{Old Photo Restoration}~\cite{wan2020bringing} + \textit{TS}~\cite{lai2018learning}: We adopt a state-of-the-art old photo restoration~\cite{wan2020bringing} algorithm for per-frame processing and generate coherent video through blind temporal smoothing~\cite{lai2018learning}.
		\item \textit{BasicVSR}~\cite{chan2021basicvsr}: An advanced RNN-based method, tailored for high-quality video super-resolution.
		\item \textit{Video Swin}~\cite{liu2021video}: A fully transformer-based architecture, which performs attention mechanisms in both spatial and temporal dimensions to capture global correlations. 
		\item \textit{DeepRemaster}~\cite{iizuka2019deepremaster} is a state-of-the-art old film restoration method using 3D convolutions, which also supports video colorization given one referenced frame. 
		\item \textit{DeOldify}~\cite{DeOldify}: An open-source tool for restoring old films.
	\end{itemize}
	
	For fair comparison, we re-train \textit{BasicVSR}~\cite{chan2021basicvsr} and \textit{Video Swin}~\cite{liu2021video} from scratch, and fine-tune \textit{DeepRemaster}~\cite{iizuka2019deepremaster} using the same training data as ours. For \textit{Old Photo Restoration}~\cite{wan2020bringing} and professional restoration tool \textit{DeOldify}~\cite{DeOldify}, we directly utilize their pre-trained models for inference.

	\begin{table}[t]
		\centering
		\small
		\begin{adjustbox}{max width=\linewidth}
			\begin{tabular}{l|cccc}
				
				\noalign{\hrule height 0.3mm} 
				\rowcolor[HTML]{F5F5F5} 
				Method       & PSNR${\uparrow}$ & SSIM${\uparrow}$ & LPIPS${\downarrow}$ & $E_{warp}$ ${\downarrow}$ \\ \hline
				Input & 19.982           & 0.699            & 0.456              & 0.0167           \\
				Old Photo+TS~\cite{wan2020bringing,lai2018learning} & 21.962           & 0.768            & 0.315               & 0.0041           \\
				BasicVSR~\cite{chan2021basicvsr}     & 23.363           & 0.808            & 0.328               & 0.0053           \\
				Video Swin~\cite{liu2021video}   & 22.758           & 0.774            & 0.319               & 0.0061           \\
				DeepRemaster~\cite{iizuka2019deepremaster} & 20.634           & 0.728            & 0.427               & 0.0066           \\
				DeOldify~\cite{DeOldify}     &    20.051              & 0.708                 &     0.436                &       0.0149           \\ \hline
				
				\rowcolor[HTML]{F7FAFE} 
				Ours         & \textbf{24.465}           & \textbf{0.840 }           & \textbf{0.192}               & \textbf{0.0019}           \\ 
				\rowcolor[HTML]{F7FAFE} 
				Ours w/o bi-direction         & 24.251           & 0.831            & 0.207               & 0.0036           \\ 
				\rowcolor[HTML]{F7FAFE} 
				Ours w/o soft mask         & 24.297           & 0.827            & 0.243               & 0.0025           \\ 
				\rowcolor[HTML]{F7FAFE} 
				Ours w/o transformer         & 24.342           & 0.830            & 0.229               & 0.0023           \\ 
				\noalign{\hrule height 0.3mm}

			\end{tabular}
		\end{adjustbox}
		\vspace{-0.3em}
		\caption{{Quantitative restoration comparisons on synthetic dataset.} Our method achieves better performance on all metrics.}
		\label{tab:restoration_compare}
		\vspace{-0.7em}
	\end{table}

	\begin{table}[t]
		\small
		\centering
		\begin{adjustbox}{max width=\linewidth}
			\begin{tabular}{l|cccc}
				
				\noalign{\hrule height 0.3mm} 
				\rowcolor[HTML]{F5F5F5} 
				Method       & PSNR${\uparrow}$ & SSIM${\uparrow}$ & LPIPS${\downarrow}$ & FID ${\downarrow}$ \\ \hline
				Input & 27.100           & 0.945            & 0.189              & 110.559           \\
				$\text{DeOldify}^{*}$~\cite{DeOldify} & 26.271           & 0.937            & 0.149               & 59.686           \\
				DeepExemplar~\cite{zhang2019deep}       & 30.064               & 0.952               & 0.091                  & 37.971               \\
				DeepRemaster~\cite{iizuka2019deepremaster} & 29.253           & 0.950            & 0.127               & 40.385           \\\hline
				
				\rowcolor[HTML]{F7FAFE} 
				Ours         & \textbf{32.838}           & \textbf{0.977}            & \textbf{0.065}             & \textbf{31.992}          \\ 
				\noalign{\hrule height 0.3mm}

			\end{tabular}
		\end{adjustbox}
		\vspace{-0.3em}
		\caption{{Quantitative colorization comparisons on REDS~\cite{Nah_2019_CVPR_Workshops_REDS} dataset.} $\text{DeOldify}^{*}$: Non-reference based video colorization. }
		\label{tab:colorization_quan}
		\vspace{-0.7em}
	\end{table}

	\begin{table}[t]
		\small
		\centering
		\setlength{\tabcolsep}{3.8mm}
		\begin{adjustbox}{max width=0.9\linewidth}
			\begin{tabular}{l|cc}
				
				\noalign{\hrule height 0.3mm}
				\rowcolor[HTML]{F5F5F5} 
				Method       & NIQE${\downarrow}$ & BRISQUE${\downarrow}$  \\ \hline
				Input & 18.9907          & 53.6776                 \\
				Old Photo+TS~\cite{wan2020bringing,lai2018learning} & 17.5110           & 48.1470                    \\
				BasicVSR~\cite{chan2021basicvsr}     & 17.6842           & 62.7381             \\
				Video Swin~\cite{liu2021video}   & 18.9462           & 52.4758              \\
				DeepRemaster~\cite{iizuka2019deepremaster} & 17.9697           & 49.9638               \\
				DeOldify~\cite{DeOldify}     &    17.9062              & 51.2813             \\ \hline
				
				\rowcolor[HTML]{F7FAFE} 
				Ours         & \textbf{15.4254}           & \textbf{42.1422}                \\ 
				
				\noalign{\hrule height 0.3mm} 
			\end{tabular}
		\end{adjustbox}
		\vspace{-0.3em}
		\caption{{Quantitative restoration comparisons on real old films.}}
		\label{tab:restoration_compare_real}
		\vspace{-1.0em}
	\end{table}
	
	\begin{figure*}[!tb]
		\setlength\tabcolsep{1.0pt}
		\centering
		\small
		\begin{tabularx}{\textwidth}{ccccc}
			\raisebox{1.8\height}{\rotatebox{90}{Input}}&
			\includegraphics[height=0.18\textwidth]{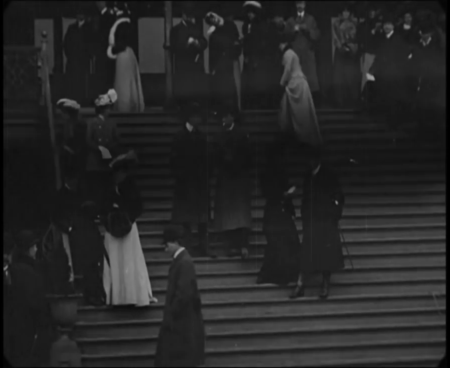}&
			\includegraphics[height=0.18\textwidth]{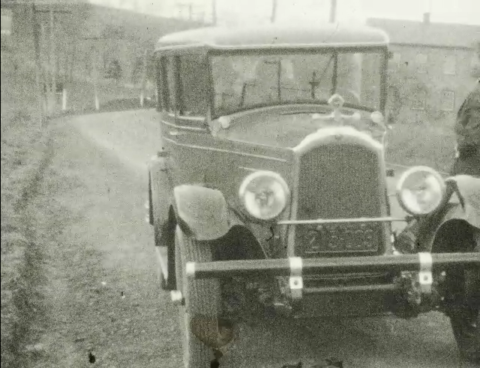}&
			\includegraphics[height=0.18\textwidth]{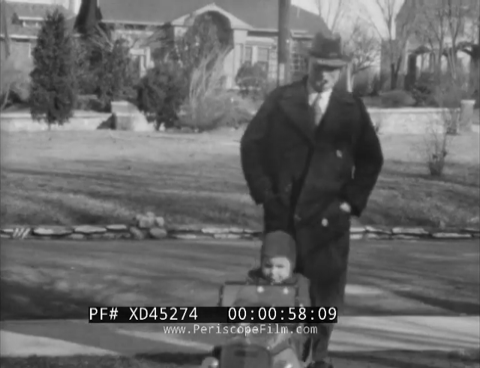}&
			\includegraphics[height=0.18\textwidth]{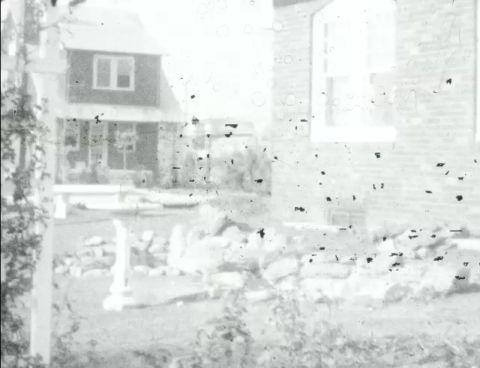}\\
			\raisebox{0.05\height}{\rotatebox{90}{Old Photo+TS~\cite{wan2020bringing,lai2018learning}}}&
			\includegraphics[height=0.18\textwidth]{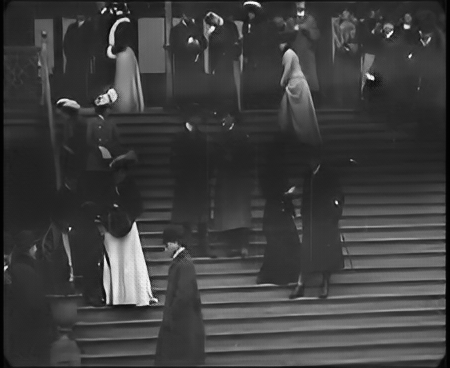}&
			\includegraphics[height=0.18\textwidth]{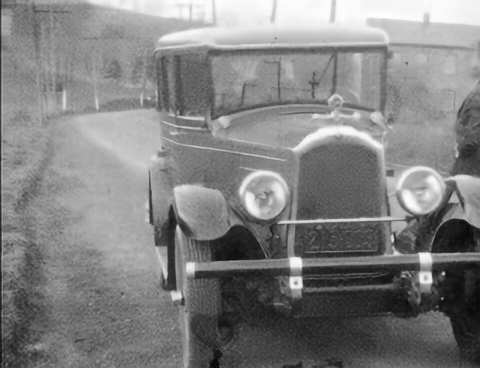}&
			\includegraphics[height=0.18\textwidth]{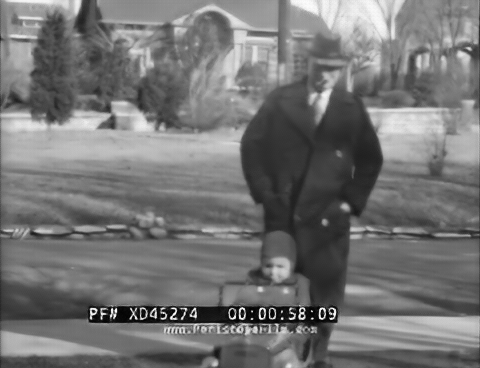}&
			\includegraphics[height=0.18\textwidth]{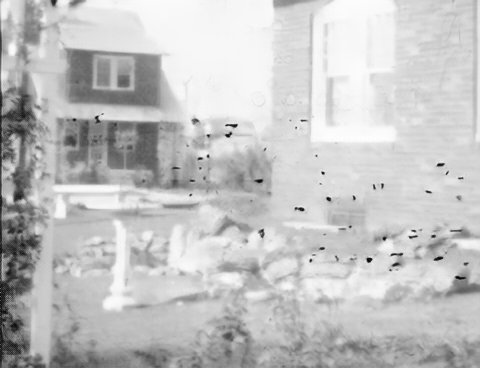}\\
			\raisebox{0.3\height}{\rotatebox{90}{VideoSwin~\cite{liu2021video}}}&
			\includegraphics[height=0.18\textwidth]{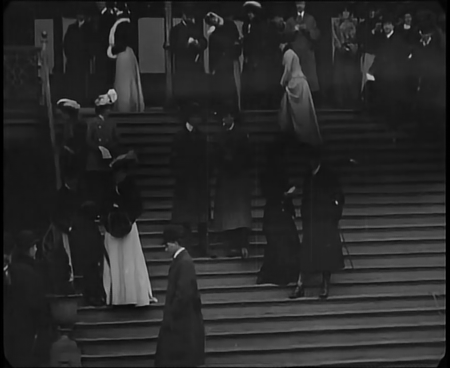}&
			\includegraphics[height=0.18\textwidth]{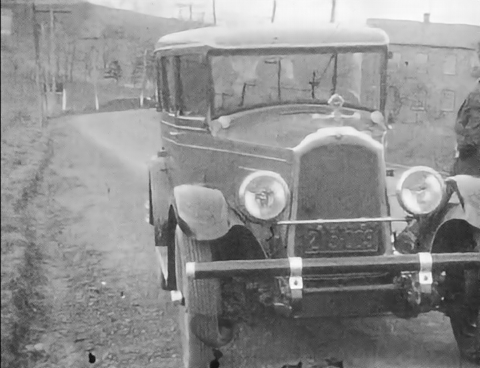}&
			\includegraphics[height=0.18\textwidth]{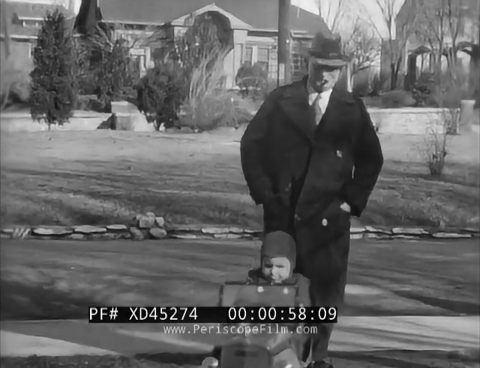}&
			\includegraphics[height=0.18\textwidth]{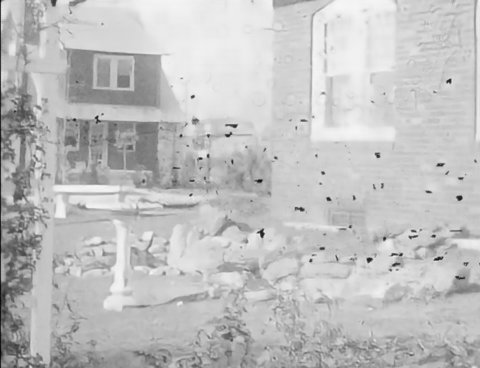}\\
			\raisebox{0.15\height}{\rotatebox{90}{DeepRemaster~\cite{iizuka2019deepremaster}}}&
			\includegraphics[height=0.18\textwidth]{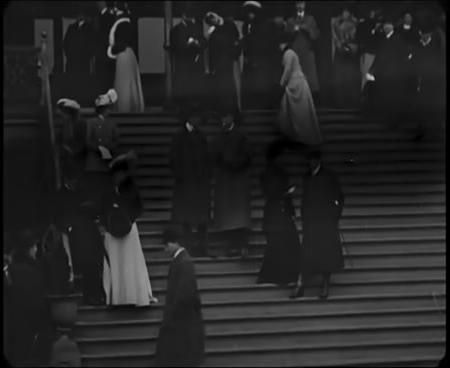}&
			\includegraphics[height=0.18\textwidth]{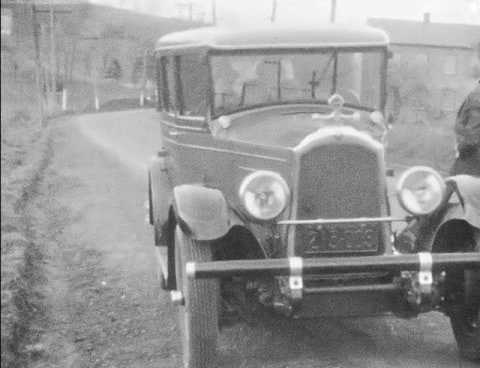}&
			\includegraphics[height=0.18\textwidth]{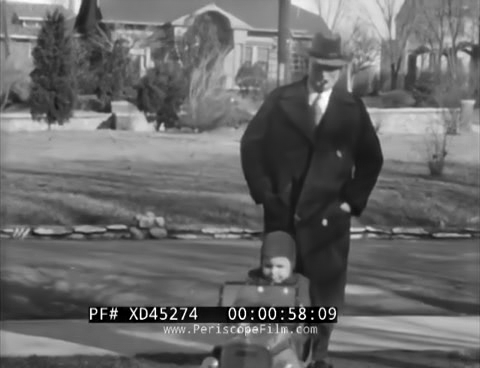}&
			\includegraphics[height=0.18\textwidth]{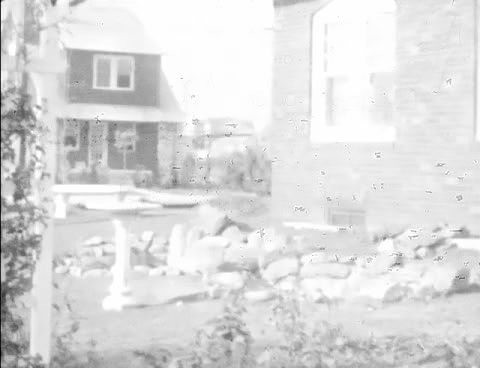}\\
			\raisebox{0.5\height}{\rotatebox{90}{DeOldify~\cite{DeOldify}}}&
			\includegraphics[height=0.18\textwidth]{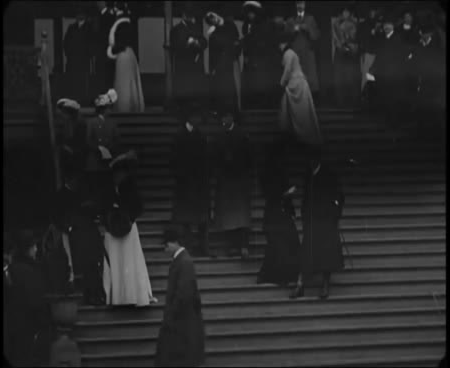}&
			\includegraphics[height=0.18\textwidth]{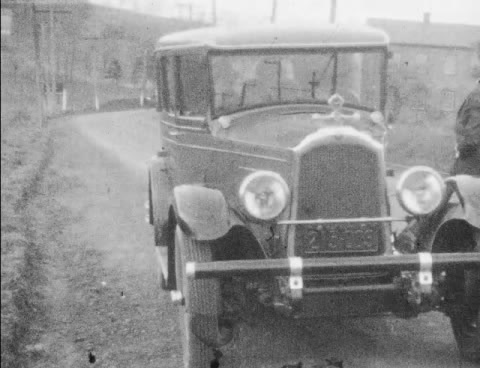}&
			\includegraphics[height=0.18\textwidth]{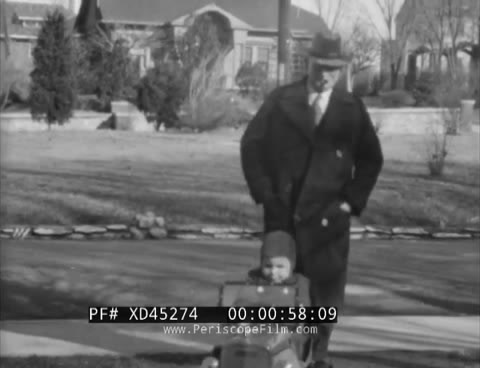}&
			\includegraphics[height=0.18\textwidth]{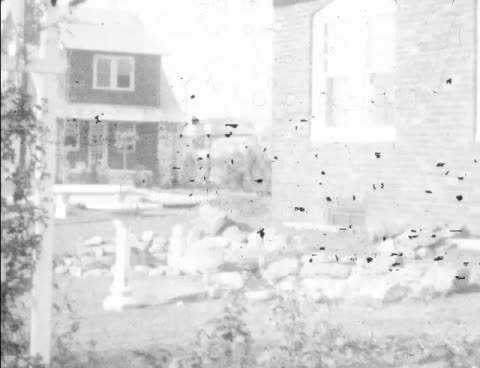}\\
			\raisebox{1.9\height}{\rotatebox{90}{Ours}}&
			\includegraphics[height=0.18\textwidth]{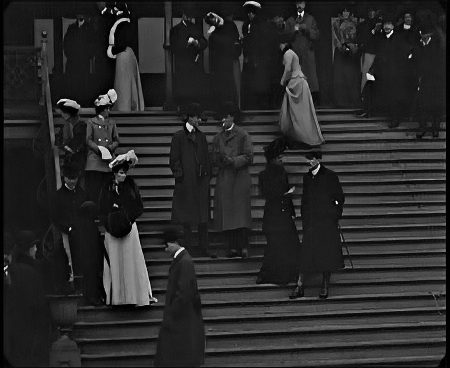}&
			\includegraphics[height=0.18\textwidth]{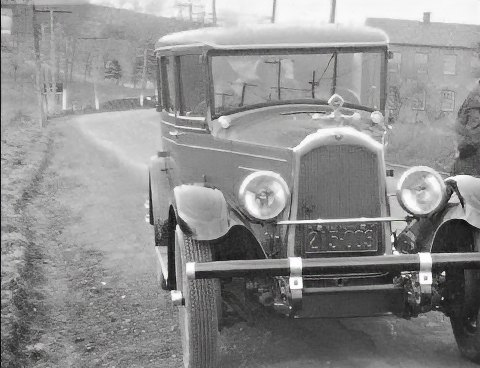}&
			\includegraphics[height=0.18\textwidth]{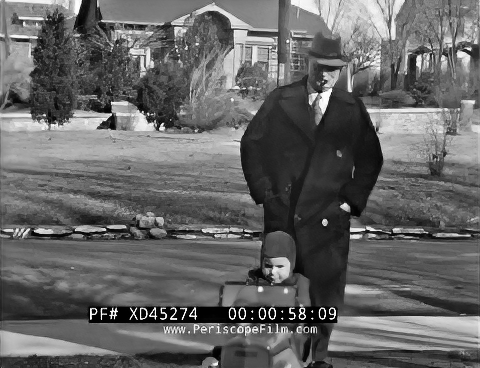}&
			\includegraphics[height=0.18\textwidth]{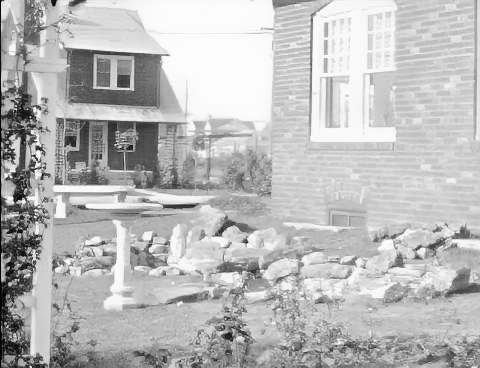}
		\end{tabularx}
		\vspace{-0.7em}
		\caption{{Qualitative restoration comparisons on real-world old films.} Our method could handle complicated degradations of old films.}
		\label{fig:comparison_realold}
		\vspace{-1.75em}
	\end{figure*}
	
	
	\def\swthree{0.2\linewidth}
	\renewcommand{\tabcolsep}{0.5pt}
	\begin{figure}
		\begin{center}
			\small
			\begin{tabular}{ccccc}
				\vspace{-0.5mm}\includegraphics[width=\swthree]{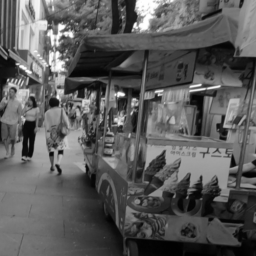}&
				\includegraphics[width=\swthree]{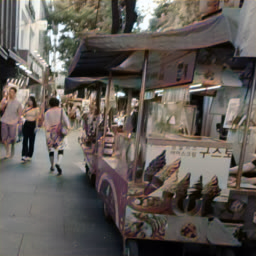}&
				\includegraphics[width=\swthree]{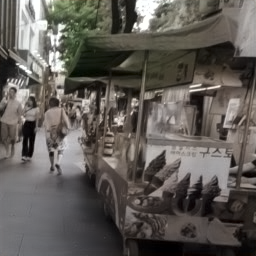}&
				\includegraphics[width=\swthree]{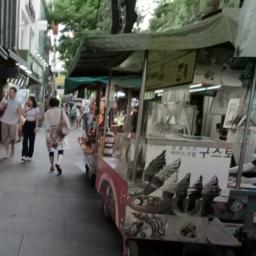}&
				\includegraphics[width=\swthree]{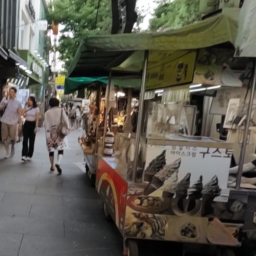}\\
				\includegraphics[width=\swthree]{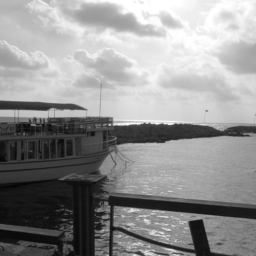}&
				\includegraphics[width=\swthree]{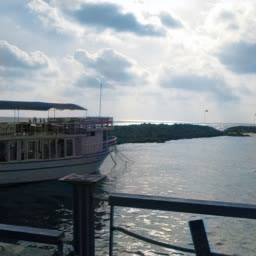}&
				\includegraphics[width=\swthree]{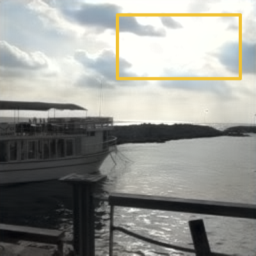}&
				\includegraphics[width=\swthree]{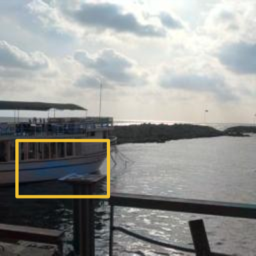}&
				\includegraphics[width=\swthree]{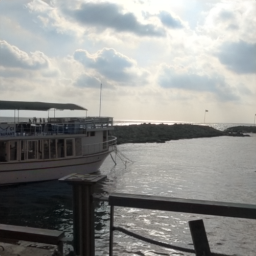}\\
				Input & \cite{DeOldify}$^{*}$~  & \cite{iizuka2019deepremaster}  & \cite{zhang2019deep} &  Ours\\
			\end{tabular}
		\end{center}
		\vspace{-1.4em}
		\caption{{Qualitative video colorization comparisons.} \cite{DeOldify}$^{*}$: Colorization without reference frame.}
		\vspace{-1.5em}
		\label{fig:colorization_quali}
	\end{figure}
	
	\noindent\textbf{Evaluation Metrics} ~~ 1) We employ peak signal-to-noise ratio (PSNR) and the structural similarity index (SSIM) to measure the low-level discrepancy between the restored output and the ground truth for synthetic data. 2) To better match the judgment of human perception, we also calculate the learned perceptual image patch similarity (LPIPS)~\cite{zhang2018perceptual}. 3) 
	We involve the temporal warping error $E_{warp}$ following~\cite{chen2017coherent,lai2018learning} to measure the temporal consistency of the restored frames. 4) Since the restoration ground truth for old films is unavailable, two non-reference frame quality assessment metrics NIQE~\cite{NIQE} and BRISQUE~\cite{BRISQUE} are used to evaluate the algorithm performance on real-world data. 5) For the video colorization task, we also adopt Fr\'echet Inception Distance (FID)~\cite{FID} to measure the semantic discrepancy between the colorized output and the natural colorful frames. 

	\subsection{Results}
	\noindent\textbf{Quantitative Comparisons} ~~ We report quantitative results on both synthetic dataset and real-world old films. We create the synthetic data by blending random degradations with clean frames of DAVIS~\cite{pont20172017} dataset, which contains large camera motions and scene diversities. As shown in Table.~\ref{tab:restoration_compare}, among these baselines, BasicVSR~\cite{chan2021basicvsr} obtains good PSNR and SSIM performances. However, because video 
	SR mainly considers unstructured degradation, the visual restored results become over-smooth while directly training with various defects, which is also indicated by the descent of LPIPS. Moreover, although per-frame old photo restoration~\cite{wan2020bringing} followed by temporal smoothing~\cite{lai2018learning} pipeline has acceptable results on LPIPS and $E_{warp}$, the PSNR and SSIM inevitably decrease without leveraging the temporal clues well. By contrast, our method outperforms the baselines on all metrics. The same conclusion still holds on real-world experiments, where we collect 63 old films from the internet for evaluation. In Table.~\ref{tab:restoration_compare_real}, the NIQE and BRISQUE results of our method are significantly better than others, which demonstrate the powerful ability of RTN to restore old films.

	\noindent\textbf{Qualitative Comparisons} ~~ We further conduct qualitative comparisons in Figure.~\ref{fig:comparison_realold}. As we can see, the old photo restoration method~\cite{wan2020bringing} only resolves a small proportion of the structured degradations since this method is designed for image restoration and does not leverage temporal information. Besides, DeOldify~\cite{DeOldify} could remove some noise but leave many other degradations unsolved. VideoSwin~\cite{liu2021video} recovers some textures through exploring the spatial-temporal correlations but struggles to remove the existing contaminants well. DeepRemaster~\cite{iizuka2019deepremaster}, as the state-of-the-art old film restoration method, is capable of handling slight dust and scratches. Nonetheless, their method could not render reasonable contents while meeting severe abrasion like the first and the last columns of Figure.~\ref{fig:comparison_realold}, meanwhile failing to produce sharp textures given blurry frames. With the elaborately designed modules, the restored videos by our RTN effectively overcome these issues. Please refer to the supplementary material and video demo for more comparisons.

	\noindent\textbf{Colorization Comparisons} ~~ In this section, we show the video colorization comparisons with other baselines. The gray version of the subset of REDS~\cite{Nah_2019_CVPR_Workshops_REDS} is adopted as the test set. For each video, we predict the colors of the first 50 frames by taking the 100th frame as the colorization reference. We choose two state-of-the-art reference-based methods DeepExemplar~\cite{zhang2019deep} and DeepRemaster~\cite{iizuka2019deepremaster}, and one old film colorization tool DeOldify~\cite{DeOldify} as baselines. Quantitatively, although DeepRemaster~\cite{iizuka2019deepremaster} leverage the reference image, perceptual metrics like LPIPS and FID in Table.~\ref{tab:colorization_quan} are not satisfactory. DeepExemplar demonstrates better performance compared with others, but color bleeding frequently appears like in Figure.~\ref{fig:colorization_quali}, which hurts the colorization quality a lot. Although our method is not specifically designed for video colorization, the combination of spatial transformer and temporal RNN solves this task well.
	
	\subsection{Ablation Study}
	
	\noindent\textbf{Temporal Bi-directional RNN} ~~ We first consider changing the bi-directional RNNs into a unidirectional setting. To ensure the fairness of this setting, we further increase the parameter number of other networks to match the bi-directional setting. Even so, the temporal consistency still vastly downgrades as shown in Table.~\ref{tab:restoration_compare}, which demonstrates the importance of perceiving both past and future temporal information while restoring the flicker artifact of old films. 
	
	\noindent\textbf{Learnable Guided Mask} ~~ Next, we replace the learnable soft mask with a direct channel-wise concatenation operation for the aggregation of the previous hidden state and the current frame feature to show its significance. Without such crucial spatial clues, the network could not distinguish the original content and contaminant artifacts well, as shown in Figure.~\ref{fig:ablation}, thereby failing to restore the structured degradations, which is consistent with the performance drop on the SSIM metric in Table.~\ref{tab:restoration_compare}. Moreover, we also visualize the learned soft mask of real-world old films in Figure.~\ref{fig:mask_show}. Although this structured degradation is never seen in the training and occupies a large portion, the predicted mask is very accurate and effective to advance restoration.
	
	\noindent\textbf{Spatial Transformer} ~~ Last but not least, we employ a CNN-based architecture instead of transformer layers to perform the spatial restoration. In this setting, the training becomes very unstable due to the underlying error of flow prediction and limited receptive field of convolutions. Therefore, we pick the best checkpoint ahead of the occurrence of gradient explosion for evaluation. Compared with the baseline, which uses a unidirectional network and transformer, although the low-level metrics like PSNR and SSIM maintain similar performance, the perceptual score, \ie LPIPS, drops a lot due to defective spatial restoration. Besides, we also show its qualitative result in  Figure.~\ref{fig:ablation}, where the scratches are not handled well without the help of the spatial transformer even though this baseline is contaminant-aware. Other advanced spatial transformers like CSwin\cite{dong2022cswin} are worthy to be explored in the future.

	\def\swthreee{0.33\linewidth}
	\begin{figure}
		\renewcommand{\tabcolsep}{0.5pt}
		\begin{center}
			\small
			\begin{tabular}{ccc}
				\vspace{-0.5mm}\includegraphics[width=\swthreee]{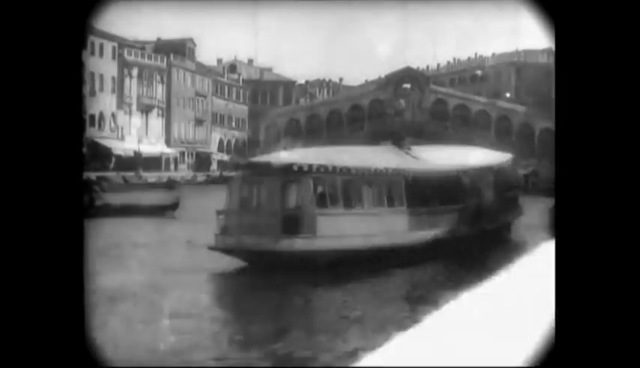}&
				\includegraphics[width=\swthreee]{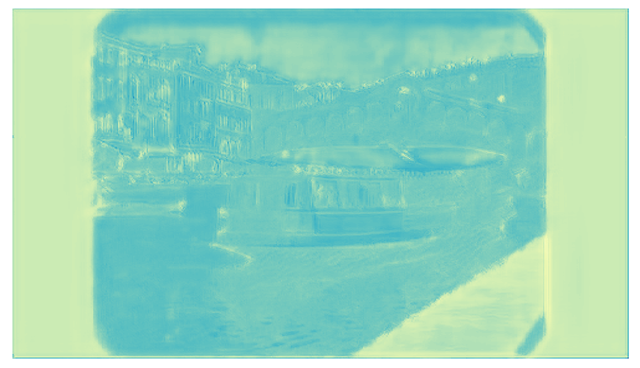}&
				\includegraphics[width=\swthreee]{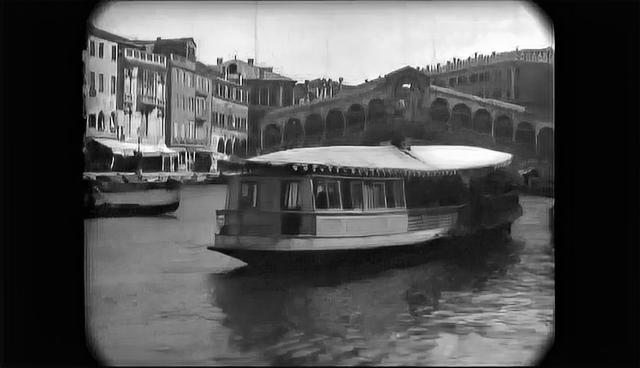} \\
				Input & Mask  & Output  \\
			\end{tabular}
		\end{center}
		\vspace{-1.4em}
		\caption{{Visualization of the learned guidance mask.} The soft mask could effectively help resolve the structured degradations.}
		\label{fig:mask_show}
		\vspace{-1.0em}
	\end{figure}

	\def\swthreee{0.33\linewidth}
	\begin{figure}
		\renewcommand{\tabcolsep}{0.5pt}
		\begin{center}
			\small
			\begin{tabular}{ccc}
				\vspace{-0.5mm}\includegraphics[width=\swthreee]{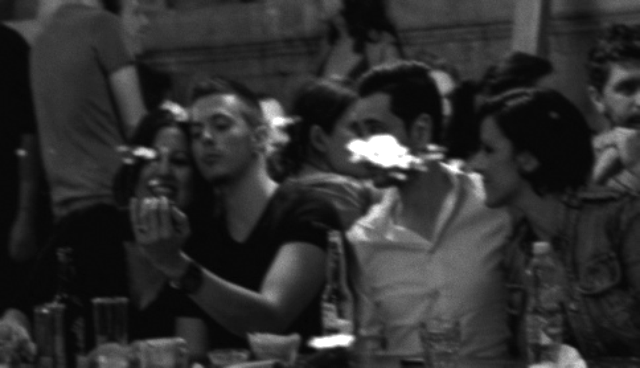}&
				\includegraphics[width=\swthreee]{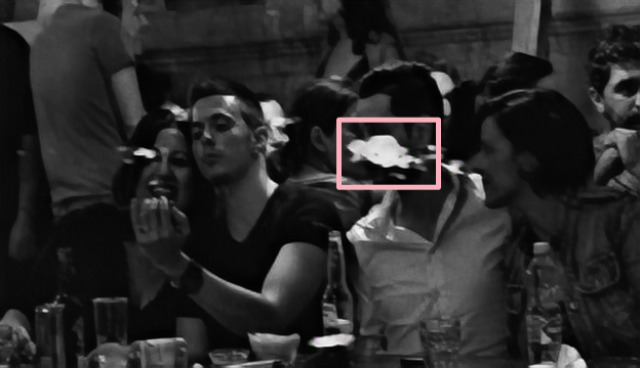}&
				\includegraphics[width=\swthreee]{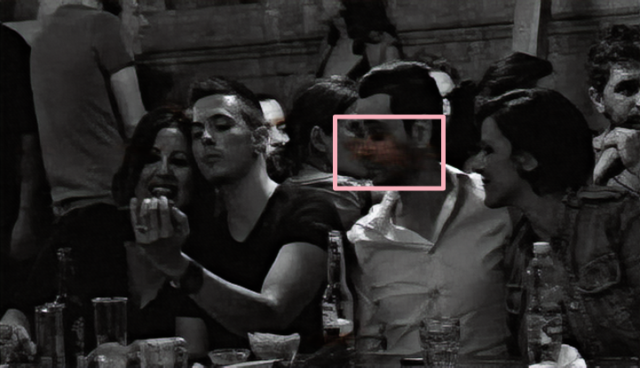}\\
				Input & w/o learnable mask  & w/o transformer  \\
				\includegraphics[width=\swthreee]{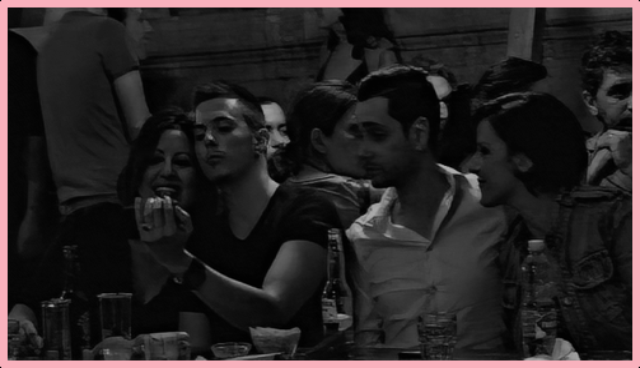}&
				\includegraphics[width=\swthreee]{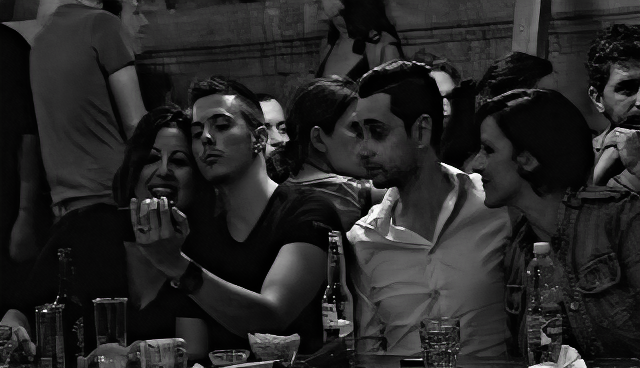}&
				\includegraphics[width=\swthreee]{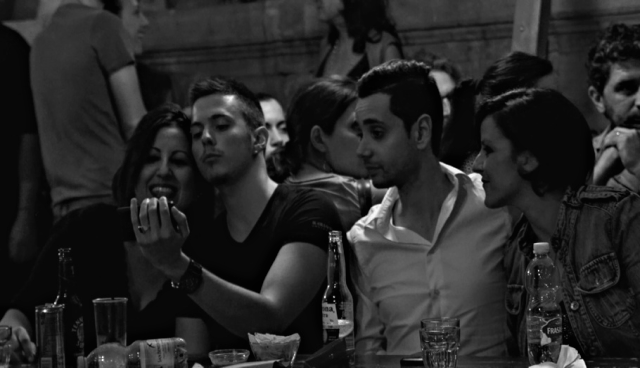}\\
				w/o bi-direction & Full model  & GT  \\
			\end{tabular}
		\end{center}
		\vspace{-1.4em}
		\caption{{Visual results of each ablation study.}}
		\label{fig:ablation}
		\vspace{-1.5em}
	\end{figure}
	
	\section{Conclusion and Limitation}
	In this paper, we present a recurrent transformer network to solve the mixed degradations of old films by leveraging the temporal modeling of recurrent neural network and the spatial modeling of transformers. Extensive visual comparisons and quantitative evaluation demonstrate that our approach performs well on both synthetic data and real old films. We also extend the RTN to achieve better reference-based video colorization compared with prior baselines. 
	
	However, there are still some limitations waiting to be resolved: a) The model may fail to distinguish the contaminant from the frame content due to their ambiguity like the black lines, which are amplified since they are wrongly recognized as the smoke of the scene; b) GAN may synthesize inadequate high-frequency details and artifacts; c) It is challenging to restore severely degraded frames with barely recognizable content. One possible solution alleviating such issues is to leverage larger scale and more diverse video data for video model pretraining \cite{wang2021bevt}. We will explore these challenges deeper in the future. 
	
	\begin{figure}[!h]
		\begin{center}
			\scriptsize
			\includegraphics[width=1.0\linewidth]{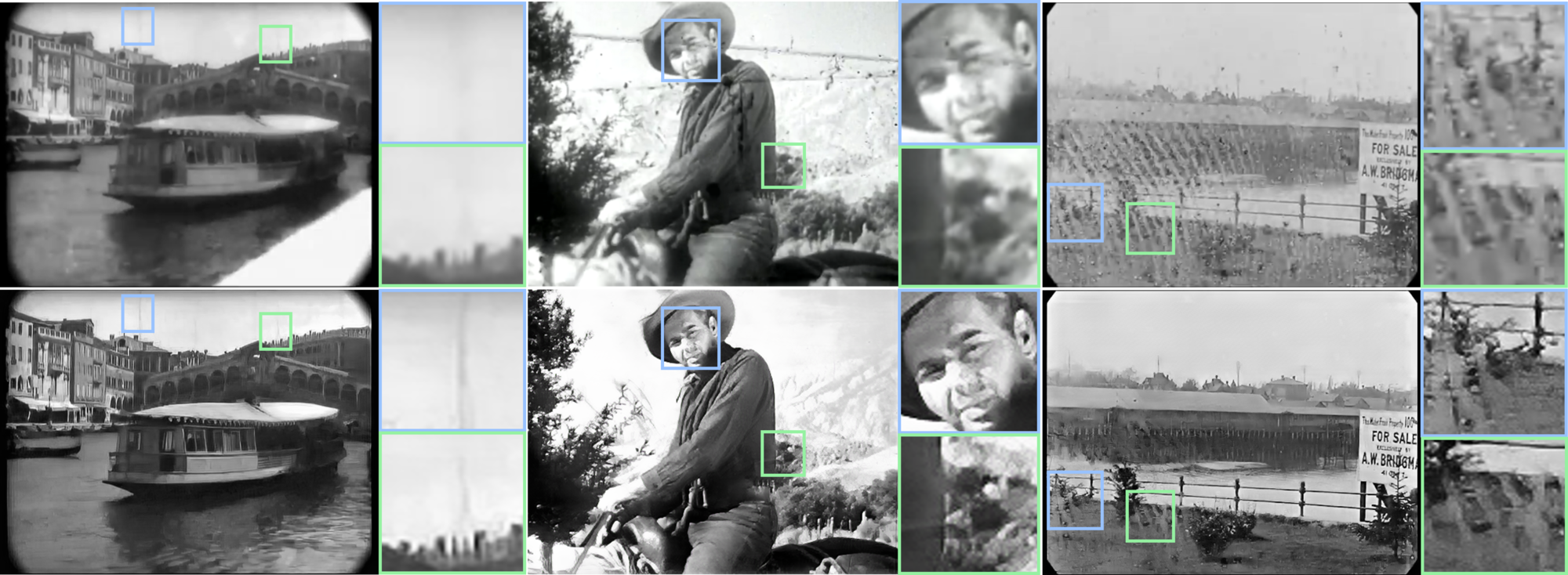}
			\renewcommand{\tabcolsep}{36pt}
			\renewcommand{\arraystretch}{0.6}
			\begin{tabular}{@{}ccc@{}}
				(a) & (b) & (c)
			\end{tabular}
			\label{fig:limitation}
		\end{center}
	\end{figure}
	\noindent\textbf{Acknowledgements: }We would like to thank anonymous reviewers for their constructive comments. This work was supported by the Hong Kong Research Grants Council (RGC) Early Career Scheme under Grant 9048148 (CityU 21209119).
	
	{\small
		\bibliographystyle{ieee_fullname}
		\bibliography{egbib}
	}
	
\end{document}